\documentclass[10pt,twocolumn,letterpaper]{article}

\usepackage{iccv}
\usepackage{times}
\usepackage{epsfig}
\usepackage{graphicx}
\usepackage{amsmath}
\usepackage{amssymb}

\usepackage{multirow}
\usepackage{bbm}
\usepackage{url}
\usepackage{booktabs,bigstrut}

\usepackage{balance}

\usepackage[pagebackref=true,breaklinks=true,letterpaper=true,colorlinks,bookmarks=false]{hyperref}

\iccvfinalcopy 

\ificcvfinal\fi

\begin{document}


\title{RPCL: A Framework for Improving Cross-Domain  Detection \\ with Auxiliary Tasks}

\author{Kai Li$^{1}$, Curtis Wigington$^{2}$, Chris Tensmeyer$^{2}$,  Vlad I. Morariu$^{2}$, \\  Handong Zhao$^{2}$,  Varun Manjunatha$^{2}$,  Nikolaos Barmpalios$^{3}$, Yun Fu$^{1}$ \\
$^1$Northeastern University, $^2$Adobe Research, $^3$Adobe Document Cloud  \\
{\tt\small \{kaili,yunfu\}@ece.neu.edu}, {\tt\small\{wigingto,tensmeye,hazhao,barmpali,morariu,vmanjuna\}@adobe.com}
}

\maketitle
\ificcvfinal\thispagestyle{empty}\fi

\begin{abstract}
Cross-Domain Detection (XDD) aims to train an object detector using labeled image from a source domain but have good performance in the target domain with only unlabeled images. Existing approaches achieve this either by aligning the feature maps or the region proposals from the two domains, or by transferring the style of source images to that of target image.  Contrasted with prior work, this paper provides a complementary solution to align domains by learning the same auxiliary tasks in both domains simultaneously. These auxiliary tasks push image from both domains towards shared spaces, which bridges the domain gap. Specifically, this paper proposes Rotation Prediction and Consistency Learning (PRCL), a framework complementing existing XDD methods for domain alignment  by leveraging the two auxiliary tasks. The first one encourages the model to extract region proposals from foreground regions by rotating an image and predicting the rotation angle from the extracted region proposals. The second task encourages the model to be robust to changes in the image space by optimizing the model to make consistent class predictions for region proposals regardless of image perturbations. Experiments show the detection performance can be consistently and significantly enhanced by applying the two proposed tasks to existing XDD methods.
\end{abstract}

\section{Introduction}
Powered by deep learning, the task of recognizing and localizing an object of interest in a scene, i.e., object detection, has been tremendously advanced in recent years \cite{girshick2015fast,girshick2014rich,ren2015faster,liu2016ssd,redmon2016you,redmon2017yolo9000,redmon2018yolov3,he2017mask}. 
While a deep learning based object detector may have impressive performance on data within the same distribution as the data the detector was trained on, its performance often drops significantly when tested on data drawn from a different distribution. This is the so-called domain shift problem.

Cross-Domain Detection (XDD) addresses the domain shift problem by jointly training a detector with unlabeled data from the domain of interest (target domain) and labeled data from an auxiliary domain (source domain) \cite{chen2018domain}. By aligning the distributions of the two domains during training, the label supervision from the source domain becomes more shareable to the target domain and hence a detector of enhanced generalizability can be obtained.  

Various approaches have been proposed to align domain distributions. The first category of approaches focus on feature alignment where images from both domains are fed to a detection network and are aligned with feature maps at different levels or  extracted region proposals \cite{zhu2019adapting,saito2019strong,chen2018domain,he2019multi,li2020cross,he2020domain}. Adversarial learning is often used to achieve this where domain classifiers try to distinguish between domains and the detection model is optimized to produce features indistinguishable between domains.
The second category of approaches 
are based on pseudo-labeling where the step of pseudo-label prediction and the step of model calibration are executed iteratively \cite{kim2019self,inoue2018cross,khodabandeh2019robust,roychowdhury2019automatic}.
A detection model, usually pretrained using the labeled source data, predicts labels on the target data. Next, the predicted labels of high confidence are selected to update the model. The third category of approaches transforms the source images to resemble the target images with generative models \cite{inoue2018cross,kim2019diversify}. 
While similar to the first category in the philosophy of alignment, these methods operate on image pixels directly instead of the feature representations. 

\begin{figure}
\centering
\includegraphics[width=0.9\linewidth]{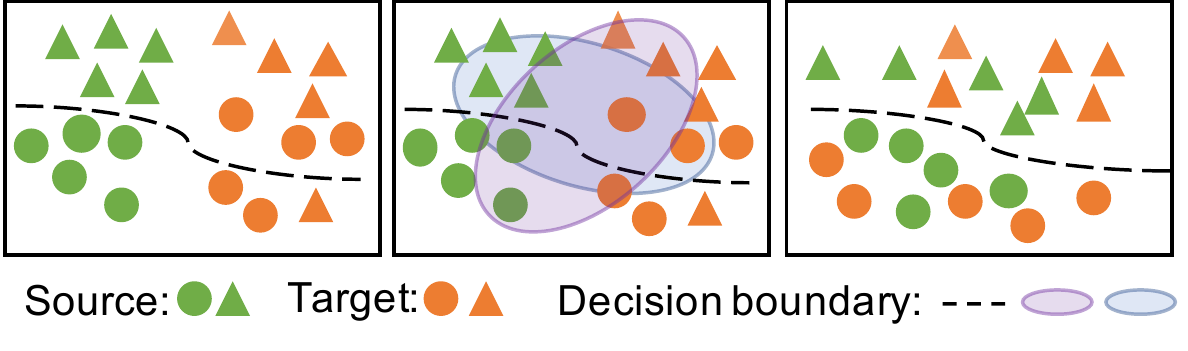}
\vspace{-0.2cm}
\caption{Illustration demonstrating the use of auxiliary tasks to align domains. Different shapes represent different classes. (Left): Without our method, the source and target domains are not well aligned and the decision boundary suited for the source domain cannot generalize well to the target domain. (Middle): The proposed auxiliary tasks treat source and target samples indiscriminately and learn shared representation spaces for all samples.
(Right): With the auxiliary tasks, the source and target domains are now closely aligned and the decision boundary generalizes well to the target domain.
}
\label{fig_concept}  
\vspace{-0.25cm}
\end{figure}

We propose techniques that are orthogonal to the three aforementioned categories and as a result complement various existing algorithms. Our main idea is to augment existing XDD models with auxiliary learning tasks that are applied on both domains simultaneously. These auxiliary tasks do not require annotated labels and thus handle source and target samples indiscriminately. The learning objectives of these task guide the model's learning differently and complement supervised learning, and thus contributing to better alignment results.
Figure \ref{fig_concept} illustrate this idea.

Specifically, we propose Rotation Prediction and Consistency Learning (RPCL), a framework that can incorporate existing XDD algorithms for enhancing adaptation performance. RPCL includes two auxiliary tasks, namely the Rotation Prediction task and Consistent Learning task that are applied to images from both domains simultaneously. For the rotation prediction task,
we rotate a given image by a random angle and then predict the angle based on features of region proposals extracted from the image. This task encourages the model to extract region proposals from the foreground regions because background regions usually lack semantics sufficient to predict the rotation angles.
The consistency learning task first perturbs a given image and then enforces the consistency of the same set of region proposals for the classification labels predicted in the original image and the augmented one. This task forces the model to be robust to changes in the image space and improve its capability of handling the domain gap.

It is worth noting that the rotation prediction task originates from self-supervised learning \cite{gidaris2018unsupervised} and the consistency learning task  from semi-supervised learning \cite{berthelot2019remixmatch,sohn2020fixmatch,xie2019unsupervised}. While we do not propose fundamentally new auxiliary tasks, we offer insights on drawing connections among seemingly distinct tasks.
We view unsupervised domain adaptation as a special case of semi-supervised learning, where the unlabeled data are drawn from a different data distribution due to the domain shift.
With this view, we can harvest the recent progress from semi-supervised learning to address the adaptation problem. Besides, we cast a unified view towards classification and detection in the region proposal level; a region proposal, once extracted from a large scene, can be viewed as a single-object image typically used for classification. 
With this framing, techniques applied on images can also be applied on region proposals, if properly adapted.  Based on these insights, we adapt effective techniques from other tasks, unifying them in a framework that can be applied to address the target problem and leading to consistently significant improvement over existing approaches. 



\section{Related Work}

\subsection{Cross-Domain Detection}
Previous work in Cross-Domain Detection (XDD) addresses the domain shift problem by aligning the features or region proposals from the source and target domains \cite{zhu2019adapting,saito2019strong,chen2018domain,he2019multi}. The alignment is often achieved by adversarial training where domain classifiers predict the domains of the pixels/images/proposals, while the detection model aims to deceive the classifiers.
One drawback of these previous methods is that the foreground and background regions are treated equally, which is undesirable as foreground regions are more semantically meaningful.
Various techniques have been proposed to emphasize the alignment of foreground regions, including learning an image classification task to regularize the model to activate regions containing the main objects \cite{xu2020exploring}; explicitly learning objectness and centerness for every pixel, and assigning weights accordingly \cite{hsu2020every}; and exploiting attention mechanism \cite{Chen2020Harmonizing}. 
Another line of approaches trains the models iteratively by generating pseudo bounding box labels for target images and updating the models with the generated pseudo-labels \cite{kim2019self,inoue2018cross,khodabandeh2019robust,roychowdhury2019automatic}. Different methods vary in how they generate the pseudo-labels or update the model. Some methods enhance the adaptation performance by improving the input images. They usually train a style-transfer model (e.g., CycleGAN \cite{zhu2017unpaired}) using images from both domains and then apply the model to translate images from the source domain as the style of the target domain \cite{inoue2018cross,kim2019diversify}. As the image style difference narrows, adapting label supervision from the source domain to the target domain becomes easier.
We address the XDD problem in a complementary way by proposing a generic framework where existing methods can be incorporated and have performance enhanced by performing two auxiliary tasks simultaneously in both domains.

\begin{figure*}
\centering
\includegraphics[width=0.9\linewidth]{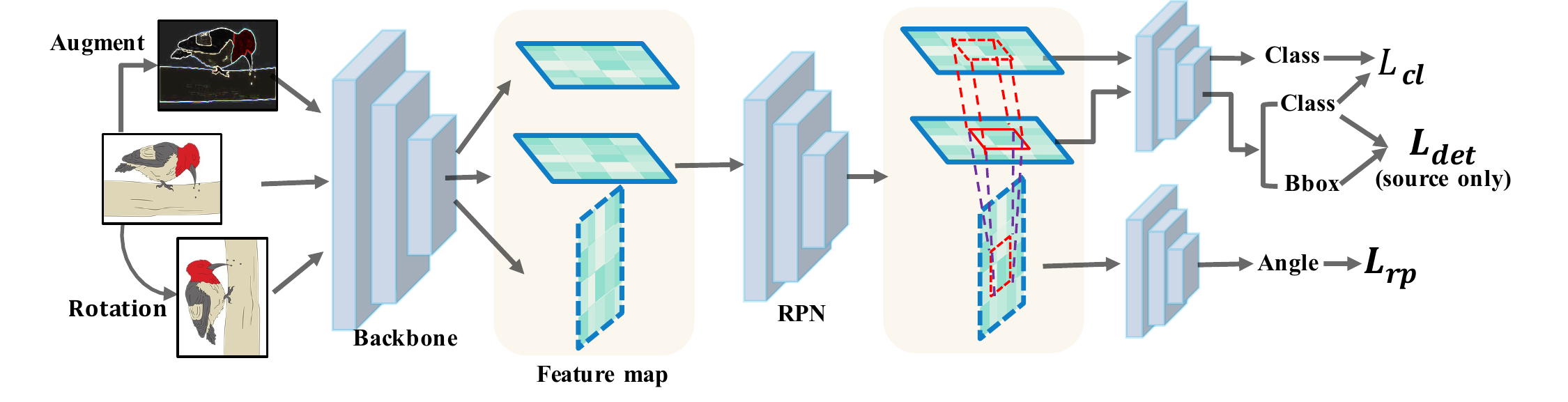}
\vspace{-0.2cm}
\caption{Illustration of the proposed framework. Our framework augments a backbone detector at training time with two additional losses for rotation prediction ($L_{rp}$) and consistency learning ($L_{cl}$) which improve performance on the target domain.}
\label{fig_framework}  
\vspace{-0.18cm}
\end{figure*}

\subsection{Self-Supervised Learning}
Self-Supervised Learning (SSL) aims to use the data itself as supervision in a pretext task where the model can learn to extract informative representations from unlabeled data. 
Early efforts focus on designing various pretext tasks including image colorization 
\cite{zhang2016colorful,larsson2017colorproxy,zhang2017split},
image rotation prediction~\cite{gidaris2018unsupervised},
spatial context prediction~\cite{doersch2015unsupervised},
solving jigsaw puzzles~\cite{noroozi2016unsupervised}, 
image inpainting~\cite{pathak2016context},
and contrastive learning~\cite{chen2020simple,he2020momentum}. 
A comparison of some of these approaches can be found in \cite{kolesnikov2019revisiting}. It shows that the simple image rotation prediction task has shown promising results. 
SSL has also been introduced to address the domain adaptive classification problem \cite{sun2019unsupervised,saito2020universal,xu2019self} where SSL is used as an auxiliary task jointly trained along with the main alignment tasks. We follow this idea but focus on the detection problem instead. Thus, rather than performing SSL task with entire images, we apply it on region proposals. To our best knowledge, this is the first use of SSL to address the XDD problem.

\subsection{Consistency Learning}
Consistency learning regularizes model predictions to be invariant to moderate changes applied to input examples. It has been a popular technique in recent semi-supervised learning literature \cite{berthelot2019mixmatch,berthelot2019remixmatch,xie2019unsupervised,sohn2020fixmatch}.
Different consistency training methods vary in how data perturbations are generated and how the consistency loss is composed. Some methods perturb images by compositing various image transformation techniques, including translation, flipping, rotation, stretching, shearing, adding noise, etc. \cite{dosovitskiy2014discriminative, cubuk2019randaugment,berthelot2019remixmatch}. MixUp \cite{zhang2017mixup}, a technique that performs linear interpolation between the samples to generate virtual samples, is used in \cite{berthelot2019mixmatch}. Learning based augmentation approaches have also been proposed, such as AutoAugment \cite{cubuk2019autoaugment} and population based augmentation \cite{ho2019population} which employ reinforcement learning to search for the most effective combinations of transformations.
Regarding the consistency loss, early works use the squared $l_2$ loss to minimize the discrepancy of the probabilities of different version of the same images \cite{sajjadi2016regularization,laine2016temporal}. The following methods replace it with the cross-entropy loss \cite{berthelot2019remixmatch,miyato2018virtual,xie2019unsupervised,sohn2020fixmatch}.
We adopt consistency learning from semi-supervised learning to address the domain shift problem for object detection. While existing methods apply the consistency constraint at the image-level, we enforce it on the region proposals.

\section{Algorithm}
We propose a general framework that can improve different existing cross-domain detectors by applying our two novel alignment techniques. 
In this section, we will first present an overview of the proposed framework and then introduce the details of the two domain alignment techniques.

\subsection{Framework Overview}
Given a labeled dataset $\mathcal{S}=\{\mathcal{X}_s, \mathcal{Y}_s\}$ from the source domain and an unlabeled target dataset $\mathcal{T}=\{\mathcal{X}_t\}$, 
Cross Domain Detection (XDD) learns an object detector under the following framework:
\begin{equation}
   L = L_{det}(\mathcal{X}_s, \mathcal{Y}_s) + \alpha L_{uda}(\mathcal{X}_s, \mathcal{X}_t), 
   \label{uda}
\end{equation}
where $\mathcal{X}_s$ and $\mathcal{X}_t$ are the images, $\mathcal{Y}_s$ denotes the labels which specify the locations and categories of the objects, and $\alpha$ is a hyper-parameter. The first term of Eq. \eqref{uda} is the standard supervised learning objective for object detection. It includes the classification objective and bounding box regression objective using labeled images from the source domain. The second term is the unsupervised domain alignment objective that aims to align the distributions of the source and target domains. It is unsupervised in the sense that it works without the need for ground truth labels. Within this framework, many effective approaches have been proposed for the unsupervised domain alignment objective. Some align the features maps of different levels, the extracted region proposals, or their combination usually by adversarial learning \cite{zhu2019adapting,saito2019strong,chen2018domain,he2019multi}. Other methods use GANs to convert images from one domain to the other so that source domain labels can be used in the target domain \cite{inoue2018cross,kim2019diversify}.

We address XDD in a way orthogonal to existing approaches. Rather than proposing another unsupervised domain alignment technique, we investigate how auxiliary tasks can help address the domain gap. Specifically, we propose a general framework that can enhance the performance of existing XDD methods by leveraging two auxiliary tasks. The two tasks are applicable to both the source and target domains and thus serve to bridge the domain gap. The first one is the region proposal based image rotation prediction task which rotates an image and predicts the image rotation angle from the region proposals extracted from the unrotated image. The second task is the consistency learning task where the model is trained to make consistent classification predictions for the same set of region proposals within an image and its strongly augmented version. Figure \ref{fig_framework} illustrates our framework.

\begin{figure}
\centering
\includegraphics[width=0.9\linewidth]{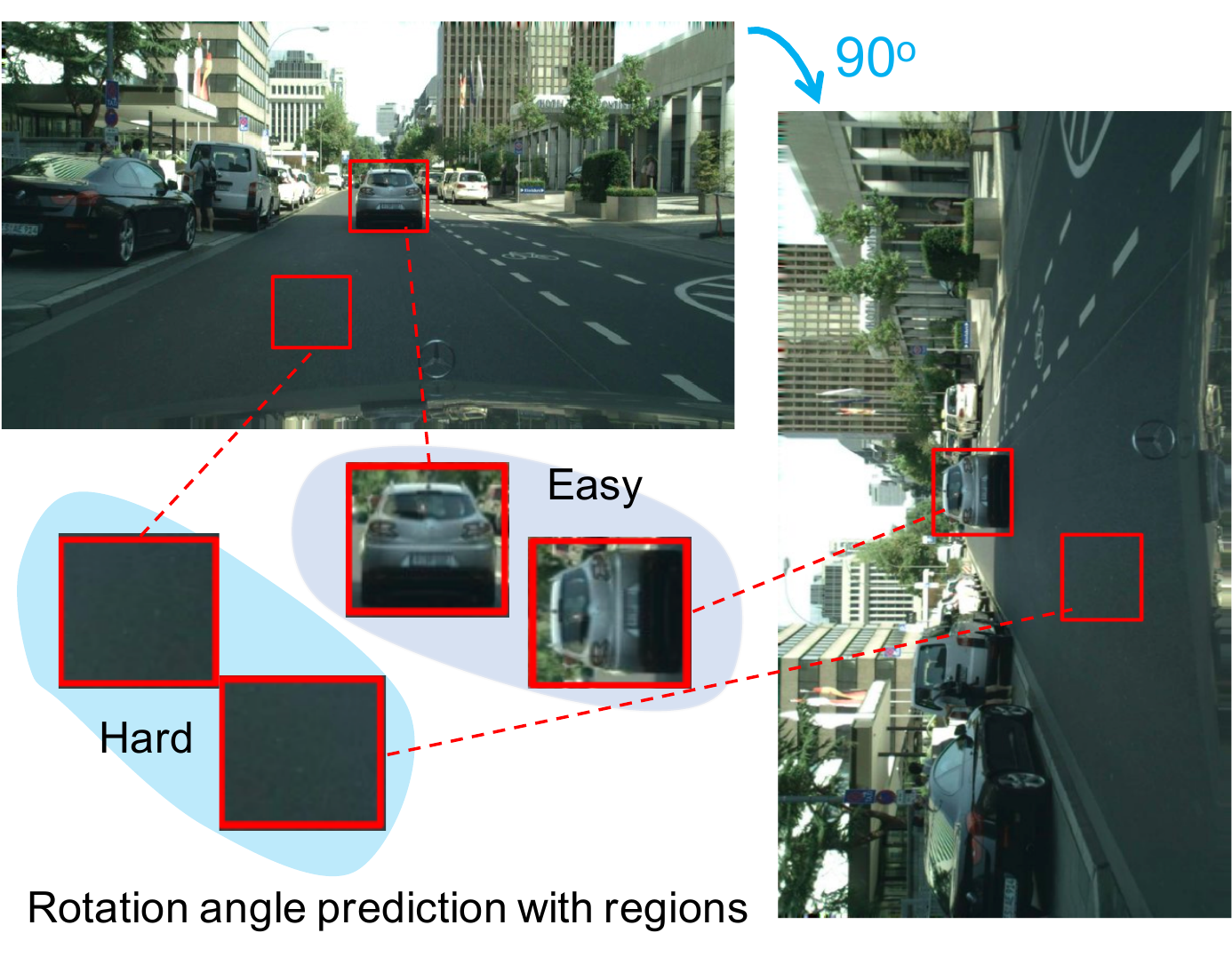}
\vspace{-0.1cm}
\caption{Predicting image rotation angle based on region proposals can help localize foreground regions.}
\label{fig_intro}  
\vspace{-0.20cm}
\end{figure}

\subsection{Proposal-Based Rotation Angle Prediction} 
Training a model to predict the rotation angle of a given image was proposed in \cite{gidaris2018unsupervised} for self-supervised learning. 
It is based on the intuition that a model can predict the rotation angle correctly if it has a deep understanding of the given image, including localization of salient objects, their orientation, the object type, etc. This inspires us to leverage this task to address the XDD problem because it does not require manually annotated labels, which suits the unsupervised domain adaptation setting well, and it helps localize salient objects and identify the object type, which is exactly the goal of object detection.

A straightforward way of exploiting this task is to learn the rotation prediction task jointly with the detection task by rotating the input image and training the model to predict the rotation angle from the feature representation of the given image.  This is how this task is utilized for the classification problem \cite{sun2019unsupervised,xiao2020self,xu2019self,su2020does}.
However, this practice is suboptimal for the detection problem because images used for detection are often much more complex, containing more salient objects in backgrounds with richer contexts. 
It may be too difficult for the model to learn a global representation for the whole image that encodes the essential information for all the salient objects.

Our insight is that classification and detection can be unified in the region proposal level: a region proposal, once extracted from a large scene, can be viewed as a single-object image typically used for classification. Based on this insight, we propose to predict the rotation angle from the region proposals. This practice has two merits.
First, it encourages the detection model to extract region proposals from the foreground since the foreground contains semantic information that are essential to predict the rotation angle. As shown in Figure \ref{fig_intro}, it is easy to tell the rotation angle from the car region, while hardly possible from the road region. Training the model to predict the rotation angle correctly encourages it to extract region proposals from foreground, which benefits for the detection task. Second, this enhances the feature alignment of foreground regions as the model will activate more on the foreground regions and thus contribute more when aligning features from the two domains.  

Formally, given a source image $\textbf{s}\in\mathcal{X}_s$, we obtain $\textbf{s}^r=Rot(x_s)$ by rotating $\textbf{s}$ with an random angle $\theta_s$ from $[0^\circ, 90^\circ, 180^\circ, 270^\circ]$.
From $\textbf{s}^r$, we extract a set of region proposals $\mathcal{R}_s$ with the 
same rotation angle $\theta_s$. 
Similarly, we can get a set of region region proposals $\mathcal{R}_t$ with the same rotation angle $\theta_t$ for every target image $\textbf{t}\in\mathcal{X}_t$.
We align the domains by applying the rotation prediction task simultaneously on the two domains. Thus, our learning objective for this task is as follows:
\begin{equation}
\begin{array}{cl}
	L_{rp}(\mathcal{X}_s, \mathcal{X}_t)= & \frac{1}{|\mathcal{X}_s||\mathcal{R}_s|}
	\sum_{\textbf{s}\sim\mathcal{X}_s}\sum_{\textbf{r}_s\sim\mathcal{R}_s} L(\textbf{r}_s, \theta_s) + \\ &
	\frac{1}{|\mathcal{X}_t||\mathcal{R}_t|}\sum_{\textbf{t}\sim\mathcal{X}_t}\sum_{\textbf{r}_t\sim\mathcal{R}_t} L(\textbf{r}_t, \theta_t),
\end{array}
	\label{ssl}
\end{equation}
where $L(\textbf{r}_s, \theta_s)$ and $L(\textbf{r}_t, \theta_t)$ are the cross-entropy losses for the source and target proposals, respectively.

\subsection{Consistency Learning} 
Consistency learning regularizes model predictions to be invariant to moderate changes applied to input examples. It has shown impressive performance for semi-supervised learning \cite{miyato2018virtual,xie2019unsupervised,sohn2020fixmatch,clark2018semi} recently. Based on the insight that unsupervised domain adaptation is a special case of semi-supervised learning where the unlabeled data is drawn from a different data distribution due to the domain shift, we propose to use consistency learning to address the XDD problem. Same as the rotation prediction task, we apply consistency learning on region proposals.

For each source image $\textbf{s}\in\mathcal{X}_s$, we apply data augmentation $\Phi$ and generate
\begin{equation}
\hat{\textbf{s}} = \Phi(\mathbf{s}).
\end{equation}
Following the previous methods \cite{sohn2020fixmatch,berthelot2019remixmatch}, we use RandAugment \cite{cubuk2019randaugment} as the data augmentation $\Phi$, which produces highly perturbed images by uniformly sampling from the image processing transformations in Python Image Library, including polarization, solarization, brightness change, color change, etc. For ease of implementation, we exclude the transformations that change the positions of pixels (e.g., flipping, rotation, etc.). This ensures $\textbf{s}$ and $\hat{\textbf{s}}$ have pixel-to-pixel correspondence for every position.  However, our framework could also work with transformations that change the position of pixels as long as the region proposals in the original image can be converted to the coordinates of the transformed image.

We extract a set of region proposals $\mathcal{R}_s$ from $\textbf{s}$ and map $\mathcal{R}_s$ directly from $\textbf{s}$ to $\hat{\textbf{s}}$, obtaining $\hat{\mathcal{R}}_s$. This ensures that every region proposal $\mathbf{r}_s\sim\mathcal{R}_s$ from $\textbf{s}$ can find the corresponding $\hat{\mathbf{r}}_s\sim\hat{\mathcal{R}}_s$ from $\hat{\textbf{s}}$ that localizes the same region in the scene. So, the pair of corresponding region proposals should be classified consistently by the classification branch of the detection model.

We enforce this consistency by optimizing the following objective function:
\begin{equation}
L^s_{cl} = \frac{1}{|\mathcal{R}_s|} \underset{{\mathbf{r}_s\sim\mathcal{R}_s, \hat{\mathbf{r}}_s\sim\hat{\mathcal{R}}_s}}{\sum} 
\big[\mathbbm{1}(\max(\textbf{p}_s)\geq\sigma) H(\textbf{p}'_s, \hat{\textbf{p}}_s)\big],
\label{self_loss}
\end{equation}
where $\textbf{p}_s$ and $\hat{\textbf{p}}_s$ are the classification probabilities of proposals $\textbf{r}_s$ and $\hat{\textbf{r}}_s$, respectively. $\textbf{p}'_s=\arg \max(\textbf{p}_s)$ returns a one-hot vector for the prediction;  $H(., .)$ is the cross-entropy of two possibility distributions; $\max(\textbf{p}_s)$  returns the highest possibility score.

In essence, we enforce consistency of the class predictions for a pair of corresponding region proposals $(\textbf{r}_s, \hat{\textbf{r}}_s)$ by computing a pseudo label from $\textbf{r}_s$ and apply the pseudo label on $\hat{\textbf{s}}$ 
by computing the standard cross-entropy loss. To mitigate the impact of incorrect pseudo labels, only the samples with confident predictions (the highest probability scores are above a threshold) are used for loss computation.

\begin{table}[t]
  \small
  \centering
  \begin{tabular}{l}
    \hline
    \noindent \textbf{Algorithm 1.} Proposed RPCL framework  \\\hline 
    \textbf{Input:} Source set $\mathcal{S}=\{\mathcal{X}_s, \mathcal{Y}_s\}$ and target set $\mathcal{T}=\{\mathcal{X}_u\}$. \\
    \textbf{Output:} Domain adaptive detector. \\\hline
    \textbf{while} not done \textbf{do}\\     
    \hspace{3mm} 1. Randomly sample $(\textbf{s}, y_s)\sim\mathcal{S}$ and $\textbf{t}\sim\mathcal{T}$.  \\ 
    \hspace{3mm} 2. Rotate $\textbf{s}$ and get $(\textbf{s}^r, \theta_s)=Rot(\textbf{s})$; rotate $\textbf{t}$ and get \\
    \hspace{7mm}    $(\textbf{t}^r, \theta_t)=Rot(\textbf{t})$; augment $\textbf{s}$ and get $\hat{\textbf{s}}=\Phi(\textbf{s})$; \\
    \hspace{7mm} augment $\textbf{t}$ and get $\hat{\textbf{t}}=\Phi(\textbf{t})$. \\
    \hspace{3mm} 3. Feed-forward $(\textbf{s}, \textbf{s}^r, \hat{\textbf{s}}, \textbf{t}, \textbf{t}^r, \hat{\textbf{t}})$ to the model. \\
    \hspace{3mm} 3. Calculate the detection loss and unsupervised domain \\ \hspace{7mm} alignment loss in Eq. \eqref{uda} using $(\textbf{s}, y_s)$ and $\textbf{t}$. \\
    \hspace{3mm} 4. Calculate the rotation prediction loss in Eq. \eqref{ssl} \\ 
    \hspace{7mm} using $(\textbf{s}, \theta_s)$ and $(\textbf{t}, \theta_t)$. \\
    \hspace{3mm} 5. Calculate the consistency learning loss in Eq. \eqref{fm}  \\
    \hspace{7mm} using $(\textbf{s}, \hat{\textbf{s}})$ and $(\textbf{t}, \textbf{t}')$. \\
    \hspace{3mm} 6. Back-propagate the loss in Eq. \eqref{RPCL}. \\
    \textbf{end while} 
  \\ \hline 
  \vspace{0.5pt}
  \end{tabular}
  \vspace{-15pt}
\end{table}

\begin{table*} [t]
	\small
	\renewcommand{\tabcolsep}{8pt}
	\begin{center}
		\begin{tabular}{ll|cccccccc|c} \hline
			&               & person &  rider &  car &  truck &  bus &  train &  mbike &  bicycle &  mAP \\	\hline       
			\multicolumn{2}{c|}{Source only}                & 17.8 & 23.6 & 27.1 & 11.9 & 23.8 & 9.1 & 14.4 & 22.8 & 18.8 \\
			\multicolumn{2}{c|}{$\textrm{DAF}^{*}$~\cite{chen2018domain}}       & 25.0 & 31.0 & 40.5 & 22.1 & 35.3 & 20.2 & 20.0 & 27.1 & 27.6 \\
         	\multicolumn{2}{c|}{DAF~\cite{chen2018domain}}       & 31.5 & 40.9 & 43.9 & 21.4 & 34.2 & 20.2 & 27.8 & 35.4 & 31.9 \\
         	\multicolumn{2}{c|}{SWDA~\cite{saito2019strong}}      & 29.9 & 42.3 & 43.5 & 24.5 & 36.2 & 32.6 & 30.0 & 35.3 & 34.3 \\         
			\multicolumn{2}{c|}{SC-DA~\cite{zhu2019adapting}}    & 33.5 & 38.0 & 48.5 & 26.5 & 39.0 & 23.3 & 28.0 & 33.6 & 33.8 \\
			\multicolumn{2}{c|}{MAF~\cite{he2019multi}}            & 28.2 & 39.5 & 43.9 & 23.8  & 39.9 & 33.3 & 29.2 & 33.9 & 34.0 \\			
			\multicolumn{2}{c|}{DAM~\cite{kim2019diversify}}      & 30.8 & 40.5 & 44.3 & 27.2 & 38.4 & 34.5 & 28.4 & 32.2 & 34.6 \\
			\multicolumn{2}{c|}{GA-CA \cite{hsu2020every}}   & 41.9 & 38.7 & \textbf{56.7} & 22.6 & 41.5 & 26.8 & 24.6 & 35.5 & 36.0 \\      
			\multicolumn{2}{c|}{ECR-DAF~\cite{xu2020exploring}}   & 29.7 & 37.3 & 43.6 & 20.8 & 37.3 & 12.8 & 25.7 & 31.7 & 29.9 \\
			\multicolumn{2}{c|}{ECR-SWDA~\cite{xu2020exploring}} & 32.9 & 43.8 & 49.2 & 27.2 & \textbf{45.1} & 36.4 & 30.3 & 34.6 & 37.4 \\\hline
			\multirow{6}{*}{RPCL} & DAF \cite{chen2018domain} + RP   & 32.7 & 41.3 & 44.5 & 20.6 & 39.5 & 28.0 & 27.8 & 35.3 & 33.7             \\   
                               & DAF \cite{chen2018domain} + CL   & 33.8 & 43.0 & 44.7 & 24.3 & 38.3 & 10.9 & 30.5 & \textbf{39.4} & 33.1             \\   
                               & DAF \cite{chen2018domain} + RP + CL & 34.2 & \textbf{47.1} & 49.0 & 25.1 & 37.7 & 13.4 & 33.9 & 38.9 & 34.9  \\\cline{2-11}                                  
                               & SWDA \cite{saito2019strong} + RP   & 39.8 & 37.8 & 48.1 & 32.0 & 32.9 & 41.6  & 31.8 & 25.3 & 36.2\\
                               & SWDA \cite{saito2019strong} + CL   & 41.8 & 34.3 & 47.7 & 30.8 & 33.2 & 43.1  & \textbf{34.5} & 28.3 & 36.7               \\   
                               & SWDA \cite{saito2019strong} + RP + CL & \textbf{47.6} & 35.0 & 49.4 & \textbf{33.8} & 33.6 & \textbf{44.5} & 31.8 & 28.3 & \textbf{38.0}                \\\hline   
		\end{tabular}
	\end{center}
	\vspace{-5pt}
	\caption{Results of adapting \textit{Cityscapes} to \textit{Foggy Cityscapes}. ``RP'' and ``CL'' stand for the proposed rotation prediction task and the consistency learning task, respectively.
	``$\textrm{DAF}^{*}$'' indicates the results reported in the paper, while ``DAF'' represents the reimplemented results. ``Source only'' stands for training the detection model using source domain data without adaptation. The best results are in \textbf{bold}.
	}
	\label{table:city}
	\vspace{-0.2cm}
\end{table*}

We apply the same consistency learning task for every target image $\textbf{t}\in\mathcal{X}_t$ as well. So, the learning objective for the consistency learning task is as follows:
\begin{equation}
	L_{cl}(\mathcal{X}_s, \mathcal{X}_t) = \frac{1}{|\mathcal{X}_s|}\sum_{\textbf{s}\sim\mathcal{X}_s} L^s_{cl} + 
	\frac{1}{|\mathcal{X}_t|}\sum_{\textbf{t}\sim\mathcal{X}_t} L^t_{cl}.
	\label{fm}
\end{equation}

\noindent\textbf{Analysis}: There are several merits of learning the above consistency learning task for the XDD problem. First, it introduces a form of consistency regularization, enforcing the model to be insensitive to the image perturbations and hence being stronger in detecting objects for unlabeled target images. Second, we generate pseudo labels for unlabeled target data and the pseudo labels share the same label space as the labeled source data. This facilitates label propagation from the labeled source domain to the unlabeled target domain. Third, we augment images with RandAugment \cite{cubuk2019randaugment}, which applies various image processing transformations. These transformations and their combinations can model a wide range of factors that cause domain shifts. By training the detection model to be resistant with these factors, the generalizability of the model is accordingly enhanced.


\subsection{Overall Learning Objective} 
Adding the learning objectives for the two tasks upon Eq. \eqref{uda}, we achieve the learning objective of our RPCL framework as follows:
\begin{equation}
\begin{array}{cl}
   \mathcal{L} & =  \mathcal{L}_{det}(\mathcal{X}_s, \mathcal{Y}_s) + \alpha\mathcal{L}_{det}(\mathcal{X}_s, \mathcal{X}_t) + \\ & \lambda_1\mathcal{L}_{rp}(\mathcal{X}_s, \mathcal{X}_t) + \lambda_2\mathcal{L}_{cl}(\mathcal{X}_s, \mathcal{X}_t) 
   \label{RPCL}
\end{array}
\end{equation}
where $\lambda_1$ and $\lambda_2$ are the hyper-parameters.

\textbf{Algorithm 1} outlines the main steps of the proposed framework.

\begin{table*}[t]
	\setlength{\tabcolsep}{1.5pt}
	\begin{center}
		\resizebox{\textwidth}{!}{
			\begin{tabular}{ll|cccccccccccccccccccc|c}
				\hline  
				&      & aero  & bike & bird & boat & bot & bus & car & cat & chair & cow & table & dog & horse & mbike & persn & plant & sheep & sofa & train & tv & mAP \\
				\hline  
				\multicolumn{2}{c|}{Source only}   &  35.6    & 52.5          & 24.3          & 23.0          & 20.0  & 43.9  & 32.8  & 10.7  & 30.6  & 11.7  & 13.8  & 6.0   & 36.8  & 45.9  & 48.7  & 41.9  & 16.5  & 7.3   & 22.9  & 32.0  & 27.8  \\
            \multicolumn{2}{c|}{DAF \cite{chen2018domain}}
            & 26.0 & 58.3 & 24.0 & 23.0 & 28.1 & 44.5 & 29.4 & 10.4 & 32.0 & 39.0 & 17.5 & 15.9 & 31.1 & 58.2 & 49.3 & 44.0 & 19.1 & 19.0 & 30.6 & 43.0 & 32.1 \\
				\multicolumn{2}{c|}{SWDA \cite{saito2019strong}}     & 26.2          & 48.5          & 32.6          & 33.7          & 38.5  & 54.3  & 37.1  & 18.6  & 34.8  & 58.3  & 17.0  & 12.5  & 33.8  & 65.5  & 61.6  & \textbf{52.0}  & 9.3   & 24.9  & 54.1  & 49.1  & 38.1  \\
				\multicolumn{2}{c|}{HTCN \cite{Chen2020Harmonizing}}  & 33.6          & 58.9          & 34.0          & 23.4          & 45.6  & 57.0  & 39.8  & 12.0  & 39.7  & 51.3  & 21.1  & 20.1  & 39.1  & 72.8  & 63.0  & 43.1  & 19.3  & 30.1  & 50.2  & \textbf{51.8}  & 40.3  \\
				\multicolumn{2}{c|}{DDMRL \cite{kim2019diversify}}     & 25.8          & 63.2          & 24.5          & \textbf{42.4} & \textbf{47.9}  & 43.1  & 37.5  & 9.1   & 47.0  & 46.7  & 26.8  & \textbf{24.9}  & \textbf{48.1}  & 78.7  & 63.0  & 45.0  & 21.3  & \textbf{36.1}  & 52.3  & 53.4  & 41.8  \\
				\multicolumn{2}{c|}{ATF \cite{he2020domain}}   & \textbf{41.9}          & 67.0          & 27.4          & 36.4          & 41.0  & 48.5  & \textbf{42.0}  & 13.1  & 39.2  & \textbf{75.1}  & \textbf{33.4}  & 7.9   & 41.2  & 56.2  & 61.4  & 50.6  & \textbf{42.0}  & 25.0  & 53.1  & 39.1  & 42.1  \\\hline				   

         \multirow{6}{*}{RPCL} & DAF \cite{chen2018domain} + RP               & 31.0 & 57.8 & 32.4 & 25.6 & 39.0 & 54.1 & 34.5 & 9.8 & 34.0 & 31.2 & 29.0 & 9.4 & 28.5 & 62.9 & 51.1 & 43.6 & 15.8 & 27.6 & 61.9 & 43.3 & 36.1
 \\   
                               & DAF \cite{chen2018domain} + CR               & 31.4 & 45.3 & 24.3 & 24.2 & 37.8 & 51.1 & 31.1 & 15.1 & 39.1 & 44.8 & 25.3 & 5.0 & 28.7 & 74.2 & 48.1 & 48.4 & 19.9 & 27.5 & 50.3 & 46.1 & 35.9  \\   
                               & DAF \cite{chen2018domain} + RP + CL           & 39.7 & \textbf{67.6}  & 28.2 & 33.7 & 36.5 & 44.5 & 41.1 & 11.4 & \textbf{47.4} & 37.5 & 16.6 & 8.7 & 27.5 & 84.1 & 52.3 & 48.6 & 17.7 & 28.2 & 47.1 & 50.4 & 38.4     \\\cline{2-23}                                  
                               & SWDA \cite{saito2019strong} + RP       & 29.9 & 57.5 & \textbf{37.3} & 26.9 & 47.2 & 51.7 & 40.4 & \textbf{18.1} & 43.2 & 52.2 & 14.9 & 24.6 & 39.0 & 82.4 & 68.9 & 46.1 & 24.6 & 35.8 & 52.8 &  47.3 & 42.0 \\
                               & SWDA \cite{saito2019strong} + CL       & 39.6 & 61.4 & 27.9 & 26.0 & 38.4 & \textbf{62.3} & 34.5 & 15.0 & 42.5 & 27.2 & 20.8 & 15.6 & 30.2 & 79.9 & 56.5  & 47.3 & 10.8 & 30.9 & \textbf{59.4} & 44.3 & 38.5         \\   
                               & SWDA \cite{saito2019strong} + RP + CL & \textbf{46.2} & 62.6 & 36.7 & 27.9 & 47.2 & 53.9 & 40.8 & 17.3 & 42.7 & 55.2 & 18.9 & 20.2 & 37.2 & \textbf{87.6} & \textbf{70.3} & 45.2 & 29.8 & 35.3 & 52.7 & 46.4 & \textbf{43.7}               \\   \hline                               
			\end{tabular}
		}
	\end{center}
		\vspace{-0.1cm}
	\caption{Results on adaptation from \textit{PASCAL VOC} to \textit{Clipart}.}
	\label{Clipart}
	\vspace{-0.2cm}
\end{table*}

\section{Experiments}

\subsection{Datasets}
Following the previous methods \cite{saito2019strong,chen2018domain}, 
we conduct experiments on the following adaptation datasets.

\noindent\textbf{\textit{Cityscape} to \textit{Foggy Cityscape}}.
The \textit{Cityscape} dataset~\cite{Cordts_2016_CVPR} consists of $3,475$ images captured by a car-mounted camera in a urban scene. Bounding boxes of 8 classes are provided, including bus, bicycle, car, bike, person, rider, train, and truck. \textit{Foggy Cityscape} is rendered from \textit{Cityscape} using the depth information and a fog mask is applied on every image \cite{sakaridis2018semantic}. This leads to strict pixel-to-pixel correspondence between every pair of images from the two datasets. Following the previous method \cite{chen2018domain,saito2019strong}, we use $2,975$ images as the training set and the rest 500 as the validation set for both datasets.
For this adaptation experiment, we use VGG16~\cite{simonyan2014very} as the backbone network and pretrain it on ImageNet~\cite{deng2009imagenet}.

\noindent\textbf{\textit{PASCAL VOC} to \textit{Clipart}}.
The \textit{PASCAL VOC} dataset \cite{everingham2010pascal} combines the training and validation sets of both PASCAL VOC 2007 and 2012, which  results in $16,551$ training images. The dataset includes ground truth annotations for 20 classes. The \textit{Clipart} dataset include $1,000$ comic images from the same 20 classes as \textit{PASCAL VOC}. Following \cite{saito2019strong}, all $1,000$ are used for both training and test. Note that during training, the ground truth annotations are not used. We use ResNet101 as the backbone and pretrained it on ImageNet.

\noindent\textbf{\textit{PASCAL VOC} to \textit{Watercolor}}.
The source dataset is also \textit{PASCAL VOC}, but annotations from only 6 classes are employed for experiments: bike, bird, car, cat, dog, and person. 
This is because \textit{Watercolor} only has annotations for these 6 classes.
\textit{Watercolor} includes $2,000$ artistic images where $1,000$ are used for training and the other $1,000$ are used for test \cite{inoue2018cross}. 
We use ResNet-101 as the backbone and pretrained it on ImageNet.


\subsection{Implementation Details}
We apply the proposed two tasks as plug-and-play components on the two different existing XDD models, DAF\footnote{We use the PyTorch reimplementation in  
https://github.com/tiancity-NJU/da-faster-rcnn-PyTorch, which gets higher mAP than that reported in the paper for the adaptation from \textit{Cityscape} to \textit{Foggy Cityscape}. We will compare with both the re-implemented one and the reported one, when available; otherwise, we will only compare with the reimplemented results.}  \cite{chen2018domain} and SWDA \cite{saito2019strong}. For fair comparison, we do not change any model-specific hyperparameters, e.g., the learning rate, the optimizer, training epochs, etc. Please refer to the original papers for the implementation details.
Specific to our RPCL framework, the implementation details are as follows.
When VGG16 is used as the backbone, the rotation prediction branch is structurally identical to the last three FC layers in the standard VGG16 network, except the output dimension of the last FC layer is 4.
When the backbone is ResNet-101, we use a lighter architecture for the rotation prediction branch to save GPU memory. 
The structure is ``Conv3 $\rightarrow$ ReLU $\rightarrow$ Conv1 $\rightarrow$ ReLU''. 
We use mean pooling over the output feature map to get a vector representation for each proposal, which is then used for rotation prediction. 
We set the hyper-parameters $\lambda_1=0.1$ and $\lambda_2=0.1$ in the overall loss function Eq. \eqref{RPCL} for all our experiments\footnote{The hyper-parameter $\alpha$ is not introduced by our framework. It varies in different XDD methods. We keep it unchanged when implementing RPCL.}. For the threshold $\sigma$ in Eq. \eqref{self_loss}, we set it as $\sigma=0.8$ for all our experiments.

\begin{table}[t]
	\setlength{\tabcolsep}{1.5pt}
	\begin{center}
		\resizebox{\linewidth}{!}{
			\begin{tabular}{ll|cccccc|c}
            \hline
				& & bike & bird & car & cat & dog & person & mAP \\
				\hline
\multicolumn{2}{c|}{Source only}  & 68.8 & 46.8 & 37.2 & 32.7 & 21.3 & 60.7 & 44.6 \\
\multicolumn{2}{c|}{DAF~\cite{chen2018domain}}  & 89.6 & 45.3 & 37.5 & 25.5 & 24.4 & 47.9 & 45.0 \\
\multicolumn{2}{c|}{WST-BSR~\cite{kim2019self}}  & 75.6 & 45.8 & 49.3 & 34.1 & 30.3 & 64.1 & 49.9 \\
\multicolumn{2}{c|}{MAF~\cite{he2019multi}}  & 73.4 & 55.7 & 46.4 & 36.8 & 28.9 & 60.8 & 50.3 \\
\multicolumn{2}{c|}{SWDA~\cite{saito2019strong}}  & 82.3 & 55.9 & 46.5 & 32.7 & 35.5 & 66.7 & 53.3 \\
\multicolumn{2}{c|}{ATF \cite{he2020domain}} & 78.8 &\textbf{59.9} & 47.9 & 41.0 & 34.8 & 66.9 & 54.9 \\\hline
         \multirow{6}{*}{RPCL} & DAF \cite{chen2018domain} + RP & 88.7 & 50.5 & 40.9 & 32.3 & 32.9 & 55.1 & 50.1                \\   
                               & DAF \cite{chen2018domain} + CL  & 76.3 & 46.4 & \textbf{57.0} & 37.2 & 26.7 & 62.7 & 51.0               \\   
                               & DAF \cite{chen2018domain} + PR + CL  & \textbf{89.2} & 53.7 & 47.2 & 42.6 & 29.2 & 64.0 & 54.3               \\\cline{2-9}                                  
                               & SWDA \cite{saito2019strong} + RP  & 79.2 & 54.9 & 46.6 & \textbf{47.4} & \textbf{44.9} & 70.4 & 57.2       \\
                               & SWDA \cite{saito2019strong} + CL  & 88.9 & 53.7 & 49.5 & 43.6 & 36.6 & 69.3 & 56.9               \\   
                               & SWDA \cite{saito2019strong} + RP + CL  & 84.3 & 57.7 & 50.1 & 44.1 & 44.7 & \textbf{73.2} & \textbf{59.0}              \\  \hline
			\end{tabular}
		}
	\end{center}
	    \vspace{-0.1cm}
\caption{Results on adpatation from \textit{PASCAL VOC} to \textit{Watercolor}.}
\label{water}
    \vspace{-0.1cm}
\end{table}

\subsection{Experimental Results}
We report the results of applying RPCL on DAF \cite{chen2018domain} and SWDA \cite{saito2019strong}.
For each of the two methods, we evaluate the performance of adding the proposed Rotation Prediction (RP) task and Consistency Learning (CL) task, both individually and combined.
By doing so, we can see the impact of each task and how they complement each other.
Apart from the two baseline methods, we also compare with some very recent XDD algorithms to show how far we have advanced the baselines towards the state-of-the-art.
It is worth noting that RPCL is orthogonal to existing XDD methods on improving the adaptation performance.
We implement RPCL on top of DAF \cite{chen2018domain} and SWDA \cite{saito2019strong} for their popularity. 
RPCL has the potential of reaching even better performance if other more recent algorithms are incorporated into the framework.

\noindent\textbf{\textit{Cityscape} to \textit{Foggy Cityscape}}.
We can see from Table \ref{table:city} that 
RPCL significantly improves the results of the two baseline methods.
It raises the mAP of SWDA from 34.3 to 36.2 with the RP task applied, to 36.7 with the CL task applied, and further to 38.0 with both tasks jointly applied. 
Similarly, DAF is promoted from 31.9 to 33.7, 33.1 and 34.9 with RP, CL, and their combination, respectively.
These results substantiate the effectiveness of RPCL on enhancing the adaptation performance, as well as the contributing role of each of the components.
With the advancement, the gaps between the baselines to the state-of-the-art have been significantly narrowed or resolved. 
RPCL lifts the performance of SWDA to a level even better than the very recent algorithm ECR-SWDA~\cite{xu2020exploring}, which convincingly validates the effectiveness.


\begin{table}
   \renewcommand{\tabcolsep}{3pt}
   \begin{center}
         \begin{tabular}{l|c|c|c}\hline               
		& SWDA   & SWDA + ImgRot   & SWDA + PropRot    \\\hline 
person     & 29.9  & 40.8 & 39.8 \\ 
rider      & 42.3  & 35.3 & 37.8 \\ 
car        & 43.5  & 47.8 & 48.1 \\ 
truck      & 24.5  & 27.9 & 32.0 \\ 
bus        & 36.2  & 32.5 & 32.9 \\ 
train      & 32.6  & 42.7 & 41.6 \\  
mbike      & 30.0  & 26.6 & 31.8 \\ 
bicycle    & 35.3  & 23.2 & 25.3 \\\hline 
mAP        & 34.3  & 34.6 & 36.2 \\\hline
         \end{tabular}
   \end{center}
    \vspace{-0.1cm}
   \caption{ Analysis of rotation prediction based on entire images (ImgRot) versus that based on region proposals (PropRot).}
   \label{tab:rotation}
    \vspace{-0.1cm}
\end{table}

\noindent\textbf{\textit{PASCAL VOC} to \textit{Clipart}}.
Table \ref{Clipart} shows the results where our approach also improves the mAP scores of SWDA and DAF on the \textit{PASCAL VOC} to \textit{Clipart} task.
RPCL enhances the performance of the two baselines with each of its two components independently and in combination.
In particular, RPCL raises SWDA from 38.1 to 43.7 for the mAP, which is even higher than the state-of-the-art result.
Different from the adaptation from $\textit{Cityscape}$ to $\textit{Foggy Cityscape}$ where the domains are similar,
there is severe domain gap between the real dataset \textit{PASCAL VOC} and comic dataset \textit{Clipart}.
We can see that RPCL can address the severe domain gap and significantly enhance the adaptation performance.

\begin{figure}
\centering
\includegraphics[width=0.8\linewidth]{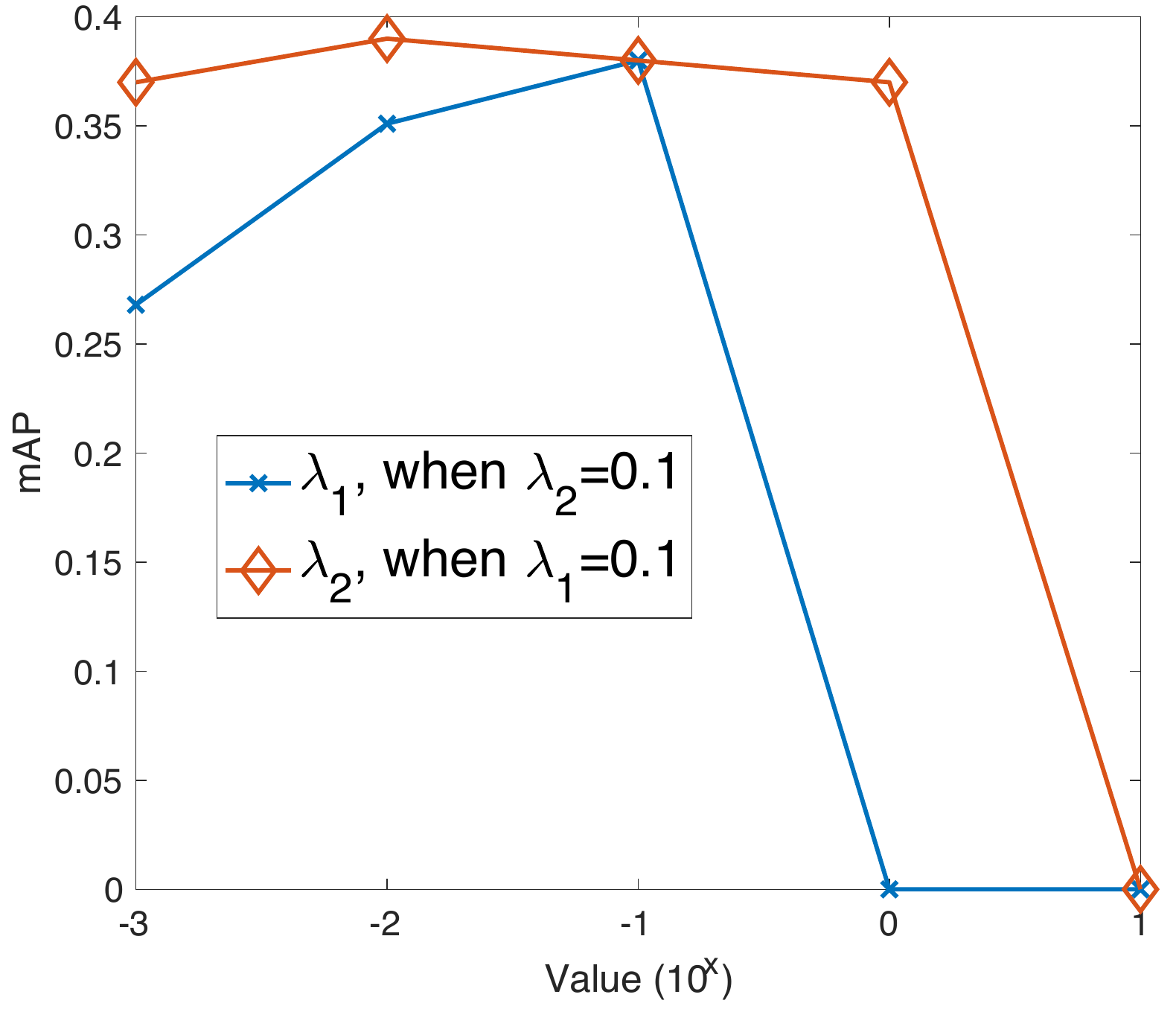}
\vspace{-0.1cm}
\caption{Parameter analysis of the proposed RPCL on top of SWDA on the adaptation from \textit{Cityscape} and \textit{Foggy Cityscape}.}
\label{fig_param}  
\vspace{-0.2cm}
\end{figure}

\begin{figure}
\centering
\includegraphics[width=0.9\linewidth]{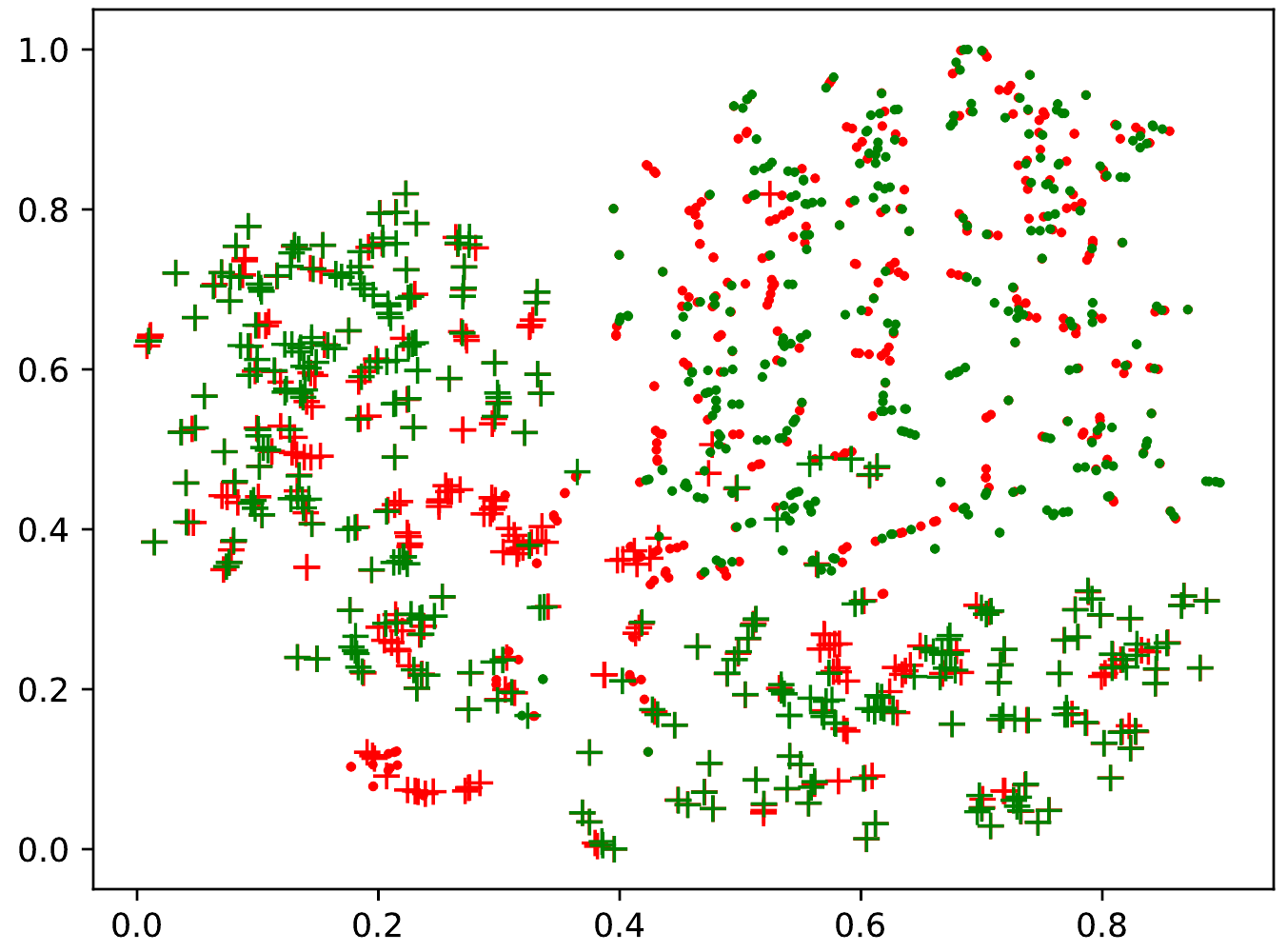}
\caption{t-SNE visualization of the ground truth box features of the car and person categories from 
\textit{Cityscape} and \textit{Foggy Cityscape}. Shapes ``+'' and ``$\circ$'' represent the two classes. Features from the same domain are drawn with the same color.}
\label{fig_tsne}  
\vspace{-0.3cm}
\end{figure}

\noindent\textbf{\textit{PASCAL VOC} to \textit{Watercolor}}.
We can further see the significant advantage of the proposed RPCL in Table \ref{water}.
RPCL raises the mAP of DAF by more than 9 points from 45.0 to 54.3, and raises the mAP of SWDA by nearly 6 points from 53.3 to 59.0, which is about 4 point improvement over the state-of-the-art performance. One possible reason for the large improvement is that there are less classes in this adaptation experiment, and thus the ambiguity between similar classes (e.g., car and bus) is not very strong. An effective alignment may lead to significant boost for the performance when the ambiguity is successfully addressed.

\begin{figure*}
\centering
\includegraphics[width=0.99\linewidth]{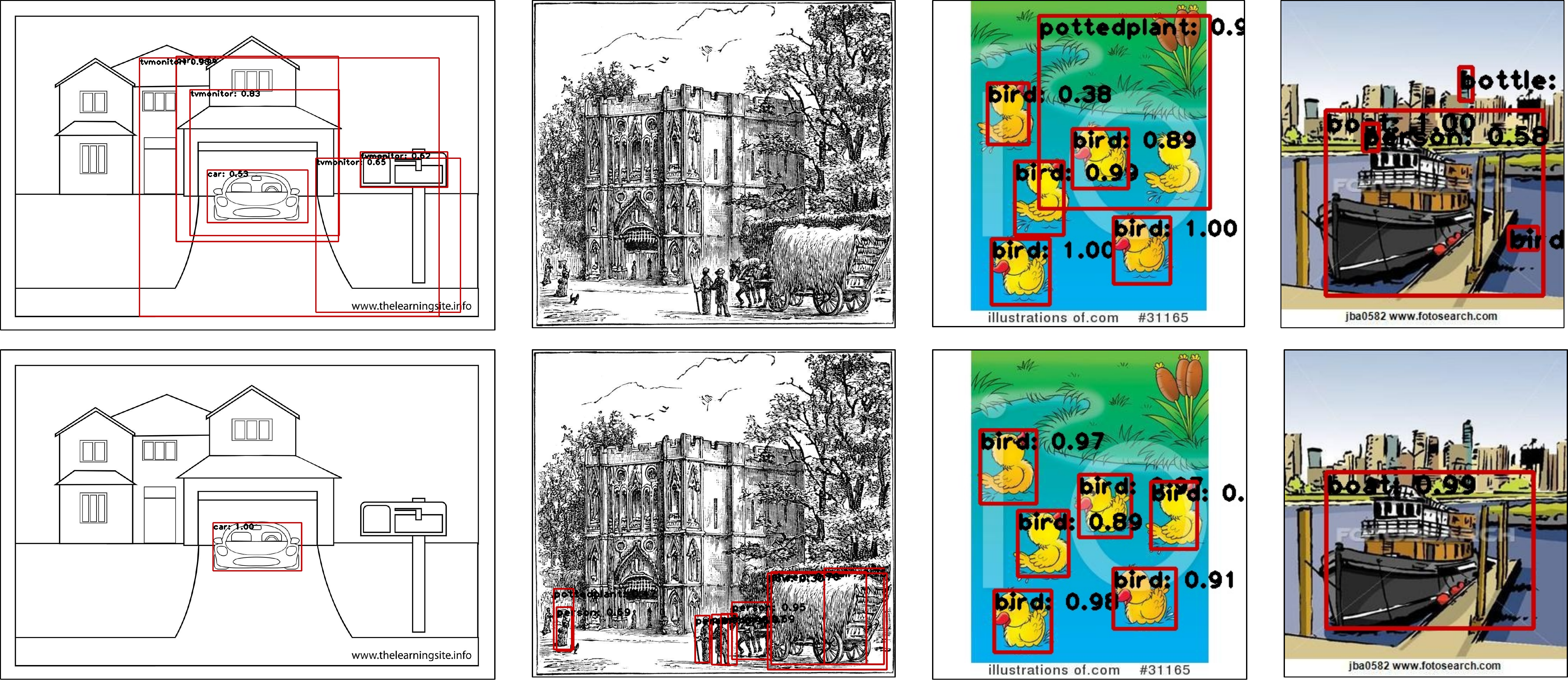}
\vspace{-0.05cm}
\caption{Detection samples. The first and second rows shows the results of SWDA \cite{saito2019strong} and RPCL on top of SWDA, respectively.
}
\label{fig_vis}  
\vspace{-0.15cm}
\end{figure*}

\subsection{Further Analysis}
\noindent\textbf{Rotation prediction from image $\textit{vs.}$ from proposals}.
One of the major differences of our rotation prediction task from the existing ones is that we predict the rotation angle based on features of region proposals extracted from an image, rather than the feature of the entire image. 
The merit is that this can encourage the model to extract region proposals from foreground regions and thus enhance detection performance. To validate this, we implement the image-based rotation prediction task and train SWDA jointly with this task. Table \ref{tab:rotation} shows the comparison on the adaption experiment from \textit{Cityscape} to \textit{Foggy Cityscape}. We can see that the image-based rotation prediction task (ImgRot) produces only a marginal improvement, which is far lower than our proposal-based rotation prediction task. This comparison verifies that the foreground regions are indeed more activated when the model is trained to extracted region proposals that facilitate to predict the rotation angle.

\noindent\textbf{Parameter analysis}.
We conducted experiments to evaluate the sensitivity of the proposed method with respect to the hyper-parameters, i.e., $\lambda_1$ and $\lambda_2$. When evaluating one parameter, we vary its value and fix the other parameter unchanged.  Figure \ref{fig_param} shows the results of RPCL-SWDA, i.e., RPCL on top of SWDA, for the adaptation from $\textit{Cityscape}$ to $\textit{Foggy Cityscape}$. We can see that RPCL is quite robust with $\lambda_2$ - the performance is stable when $\lambda_2$ varies in a wide range. RPCL is more sensitive to $\lambda_1$ and the performance drops to 0  when $\lambda_1$ is greater tnan 1. This is because the model fails to converge when the rotation prediction loss is weighted too much.

\noindent\textbf{Feature visualization}.
To qualitatively evaluate the alignment results, we plot in Figure \ref{fig_tsne} the t-SNE \cite{maaten2008visualizing} visualization of the instance features obtained by
applying RoIAlign on the ground truth instances from \textit{Cityscape} and \textit{Foggy Cityscape}. The features extracted by the RPCL-SWDA model 
for the car and person classes from both domains are shown. We can see that the source features and target features are closely aligned, while features from different classes are separated.  


\noindent\textbf{Detection samples}.
Figure \ref{fig_vis} shows some detection samples from the $\textit{Clipart}$ dataset using the RPCL-SWDA \cite{saito2019strong}.
As a comparison, we also show the detection results of SWDA on the same images. We can see from the figure that RPCL-SWDA produces fewer false negatives (real objects but not detected) and false positives (objects detected but not real). This further validates the efficacy of RPCL on improving the performance of SWDA.

\section{Conclusions}
We introduce in this paper the RPCL framework which can improve the performance of different existing Cross-Domain Detection (XDD) methods through the two introduced auxiliary tasks: the rotation prediction task and the consistency learning task.
The rotation prediction task encourages the detection model to extract region proposals from the foreground. This benefits both the detection task and domain alignment. The second task encourages the model to make smooth class predictions for region proposals when the input image has been applied with various transformations that model domain shifts. Thus, the learned model should have enhanced generalizability on the target domain. The experiments show that each of the two tasks contributes to performance gains for different XDD methods, and the tasks complement each other, pushing the baseline methods towards new state-of-the-art results.

\clearpage

{\small
\bibliographystyle{ieee_fullname}
\bibliography{egbib}
}

\end{document}